\pdfoutput=1
\documentclass[11pt]{article}

\usepackage[]{acl}

\usepackage{times}
\usepackage{latexsym}

\usepackage{graphicx,subcaption}
\usepackage[ruled,vlined]{algorithm2e}
\usepackage{algorithmic}

\usepackage{enumitem}
\setlist{nosep}

\usepackage{multirow}
\usepackage{verbdef}
\verbdef{\verbtext}{YTS}

\usepackage{amsmath}
\usepackage{amsthm}
\usepackage{amsfonts}
\usepackage{xcolor}

\definecolor{custom_pink}{RGB}{200, 20, 150}

\usepackage{array}
\newcolumntype{L}[1]{>{\raggedright\let\newline\\\arraybackslash\hspace{0pt}}m{#1}}
\newcolumntype{C}[1]{>{\centering\let\newline\\\arraybackslash\hspace{0pt}}m{#1}}
\newcolumntype{R}[1]{>{\raggedleft\let\newline\\\arraybackslash\hspace{0pt}}m{#1}}
\usepackage[T1]{fontenc}
\usepackage[utf8]{inputenc}

\usepackage{microtype}

\title{Identifying Semantically Difficult Samples to Improve Text Classification}

\author{Shashank Mujumdar \\
  IBM Research, India \\
  \And
  Stuti Mehta \\
  DA-IICT \\
  \And
  Hima Patel \\
  IBM Research, India \\
  \And
  Suman Mitra \\
  DA-IICT \\
}

\begin{document}
\maketitle
\begin{abstract}
In this paper, we investigate the effect of addressing difficult samples from a given text dataset on the downstream text classification task. We define difficult samples as being non-obvious cases for text classification by analysing them in the semantic embedding space; specifically - (i) semantically similar samples that belong to different classes and (ii) semantically dissimilar samples that belong to the same class. We propose a penalty function to measure the overall difficulty score of every sample in the dataset. We conduct exhaustive experiments on 13 standard datasets to show a consistent improvement of up to 9\% and discuss qualitative results to show effectiveness of our approach in identifying difficult samples for a text classification model.
\end{abstract}

\section{Introduction}
\label{sec:intro}
In the recent past there has been an emphasis on the assessment of quality of data for machine learning tasks \cite{jain2020overview}\cite{swayamdipta2020dataset} and a few approaches focus on assessing the training datasets. \cite{ghorbani2019data},\cite{yoon2020data} have looked at the problem of finding most valuable data instances for a chosen classifier. \cite{csaky2019improving} discuss a method for data filtering to improve the quality of data for neural conversation models in a model agnostic fashion. \cite{peinelt2019aiming} suggest profiling the datasets to find non obvious cases for semantic similarity datasets.  In this paper, we present our analysis of semantically difficult samples in the training data and their impact on the downstream models for text classification task. Table \ref{tab:airline_samples} shows examples of two types of difficult samples - (i) samples with high semantic similarity and different labels and (ii) samples with low semantic similarity and belonging to the same class. We propose an intuitive penalty function to measure the difficulty score of every sample in the dataset. We present both quantitative and qualitative results to study the effect of samples from both these categories on the performance of the downstream text classifiers. Similar to \cite{csaky2019improving}, our method is model agnostic. We present our results on $13$ standard datasets utilizing standard text classifiers and encoding schemes.

%To motivate the problem, lets look at few examples from airline tweet dataset \cite{rane2018sentiment}. Table \ref{tab:airline_samples} shows two examples, in one case, where the samples are very similar but have different class labels. In the second case, both the examples belong to the same class, however differ significantly in their sentiment. These could be due to issue of noisy labels or insufficient labels while data labeling. For example, the first sentence in the neutral class, could be part of a new class, called 'slightly negative', but the dataset in this case has only three labels. Irrespective of the source of the problem, such samples hamper the quality of the training dataset and need to be systematically analyzed before training a classifier.

\begin{table}[t]
\centering
\begin{tabular}{p{2cm}p{5cm}}
\hline
\textbf{$Class$} & \textbf{$Original \; Sample$}
\\
\hline
\hline
\verb|Neutral| & @AmericanAir Thank you
\\
\hline
\verb|Positive| & @USAirways Thank you
\\ 
\hline
% \verb|Not Spam| & Hello Brazil ðŸ˜»âœŒðŸ’“ðŸ˜»ðŸ‘ï»¿
\verb|Neutral| & @united what's a girl gotta do to get a flight name change when SHE bought one for a mean ex boyfriend and needs a girl's trip stat?!
\\
\hline
\verb|Neutral| & @AmericanAir @pbpinftworth iPhone 6 64GB (not 6 plus)
\\ 
\hline
\end{tabular}
\caption{Samples from Airline Tweets Dataset}
\label{tab:airline_samples}
\end{table}

\section{Proposed Approach}
\label{sec:approach}
% In this section we discuss our approach to compute an overall difficulty score for each sample in a given dataset.

\subsection{Difficulty of a Sample}
\label{ssec:difficulty}

\begin{figure}[t]
    \centering\includegraphics[width=1\linewidth]{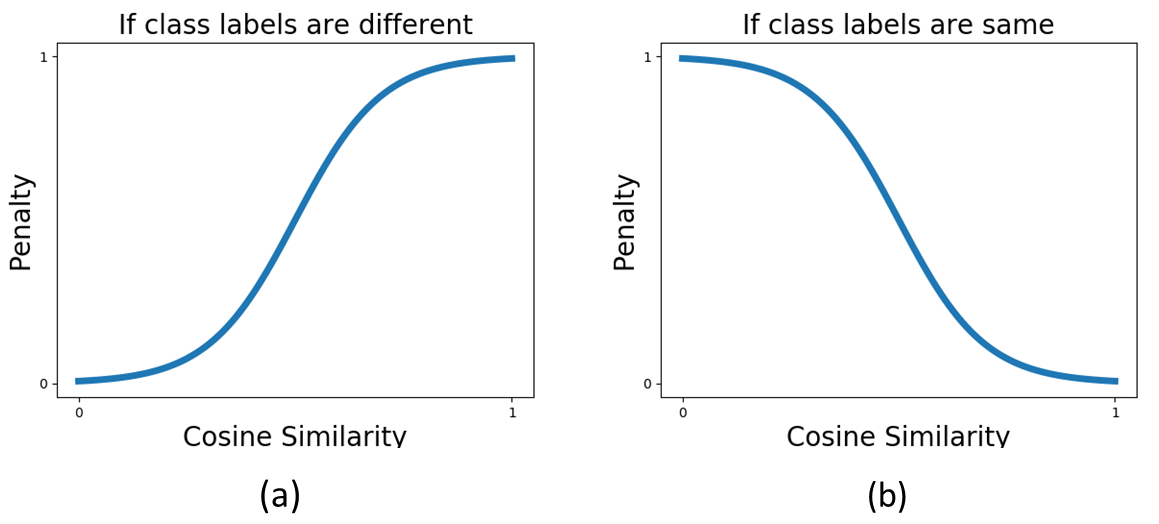}
  \caption{(a) S-penalty function (b) Z-penalty function}
  \label{fig:penalty_functions}
\end{figure}

Let $\mathcal{D}=\{(x_i,y_i)\}_{i=1}^n$ be a labelled train dataset with \textit{n} samples where $x_i$ being the input text example and $y_i$ its corresponding label. Let $e_i$ denote the encoded vector representation of the input text $x_i$. For a pair of samples in the embedding space $(e_i$, $e_j)$, we argue for the following two cases that contribute to their difficulty score,
\begin{enumerate}[leftmargin=*]
    \item \label{case:1} $e_i$ and $e_j$ are semantically similar (lie close to each other in the embedding space) but they have different labels ($y_i \neq y_j$) \hfill \textbf{[case 1]}
    \item \label{case:2} $e_i$ and $e_j$ are semantically dissimilar (lie far apart in the embedding space) but they have the same label ($y_i = y_j$) \hfill \textbf{[case 2]}
\end{enumerate}
It is intuitive as to why samples belonging to \textit{case 1} can be difficult. For \textit{case 2}, while one can expect semantic dissimilarity between a class, we look at the extreme cases, where the samples even though are part of the same class, could be referring to two different concepts within the same class. For a given sample $e_i$, we consider its pairwise relationship (PR) with all other samples in the dataset to compute its overall difficulty score. However, the intention is to only penalise the sample pair if it meets the above criterion (case \ref{case:1}\&\ref{case:2}). Thus, we introduce penalty functions (see Figure \ref{fig:penalty_functions}) that consider the PR and corresponding labels to output a penalty score for the input pair $(e_i$, $e_j)$.

\subsection{Penalty Function}
\label{ssec:penalty}

For a sample pair $(e_i$, $e_j)$, the PR is captured by the cosine similarity $cos(e_i, e_j) = \frac {e_i \cdot e_j}{||e_i|| ||e_j||}$. Instead of determining a threshold on the $cos_{sim}$ to identify if the samples are similar or dissimilar, we employ a sigmoid function to assign the penalty scores. Specifically, we use an \textit{s-shaped} sigmoid $S(x) = \frac {1}{1 + e^{(a -bx)}}$ for case \ref{case:1} and a \textit{z-shaped} sigmoid $Z(x) = \frac {1}{1 + e^{-(a -bx)}}$ for case \ref{case:2} where $a,b > 0$ and $x$ is the $cos_{sim}$. Thus, if the samples belong to different classes but have a high $cos_{sim}$, utilizing $S(x)$, a high penalty value is assigned and vice-versa. Similarly, if the samples belong to the same class but have high $cos_{sim}$, utilizing $Z(x)$, a low penalty value is assigned and vice-versa.

% For a pair of samples $(e_i$, $e_j)$, the pairwise relationship is captured by the cosine similarity $cos(e_i, e_j) = \frac {e_i \cdot e_j}{||e_i|| ||e_j||}$ between the two samples. For a given sample pair, instead of determining a threshold on the cosine similarity to identify if the samples are similar or dissimilar, we employ a sigmoid function to assign the penalty scores. Specifically, we use an \textit{s-shaped} sigmoid $S(x) = \frac {1}{1 + e^{-\alpha x}}$ for case \ref{case:1} and a \textit{z-shaped} sigmoid $Z(x) = \frac {1}{1 + e^{\alpha x}}$ for case \ref{case:2} where $\alpha > 0$ and $x$ is the cosine similarity between the sample pair, as show in Figure \ref{fig:penalty_functions}. Thus, for a given sample pair if the samples belong to different classes but have a high cosine similarity, utilizing $S(x)$, a high penalty value is assigned and vice-versa. Similarly, if the samples belong to the same class but have high cosine similarity, utilizing $Z(x)$, a low penalty value is assigned and vice-versa.

\subsection{Identifying Difficult Samples}
\label{ssec:identify_difficult}

For each sample in the dataset, a cumulative penalty score is computed by summing the pairwise penalty scores with all the other samples in the dataset. The samples in the dataset are sorted in descending order w.r.t. the cumulative penalty scores and the top $k\%$ samples are labelled as difficult. The overall approach is summarized in Algorithm \ref{algo:approach}.

\begin{algorithm}[h]
\caption{Identify Difficult Samples}
\label{algo:approach}
\SetKwInOut{Input}{Input}
\SetKwInOut{Output}{Output}
\Input{Labelled Text Dataset $\mathcal{D}$}
\Output{Difficult Samples} 
% $[\mathcal{E}] \leftarrow$ Matrix of text embeddings for each sample in the dataset.\\
% $[CP] \leftarrow$ List of cumulative penalty scores for each sample in the dataset.\\
$[\mathcal{E}] \leftarrow$ Text embeddings for each sample.\\
\begin{algorithmic}[1]
    \FOR{$i=1,2,\ldots \mathcal{|E|}$}
        \STATE $cp_i = 0$
        \FOR{$j=1,2,\ldots \mathcal{|E|}, j \neq i$}
            % \STATE Compute Pairwise Similarity
            \STATE $x = cos(e_i, e_j) = \frac {e_i \cdot e_j}{||e_i|| \cdot ||e_j||}$
            \STATE Compute Pairwise Penalty
            \IF {$y_i \neq y_j$, $y_i$ is label for $e_i$}
                \STATE $S(x) = \frac {1}{1 + e^{(a -bx)}}$
                \STATE $cp_i \mathrel{+}= S(x)$
            \ELSIF{$y_i == y_j$}
                \STATE $Z(x) = \frac {1}{1 + e^{-(a-bx)}}$
                \STATE $cp_i \mathrel{+}= Z(x)$
            \ENDIF
        \ENDFOR
        \STATE Store cumulative penalty for each sample
        \STATE $CP[i] = cp_i$
    \ENDFOR
    \STATE Sort $CP$ in descending order
    \STATE Return top $k\%$ samples from $\mathcal{D}$ using penalty scores from $CP$
\end{algorithmic}
\end{algorithm}

\section{Datasets and Experiments}
\label{sec:datasets_exp}

\subsection{Datasets}
\label{ssec:datasets}

We identify 13 standard datasets used for text classification from prior art literature as shown in Table \ref{tab:results} and use the standard split for train, validation, and test sets when available. For AT \cite{rane2018sentiment}, MR \cite{pang2002thumbs} and SE \cite{nakov2019semeval} a 10\% split from training set has been used as a validation set.

\begin{table*}[t]
\centering
% \begin{tabular}{p{1.4cm}p{1.4cm}p{1.6cm}p{1.4cm}}
\begin{tabular}{cccccccc}
\hline
\hline
\textbf{$Dataset$}&\textbf{$Source$}& \textbf{$Classes$}& \textbf{$Size$} & \textbf{$F1_{0\%}$}  &  \textbf{$F1_{1\%}$} &  \textbf{$F1_{5\%}$} &\textbf{$F1_{Best}\;(k\%)$} \\
\hline
\hline
\verb|AT|  & \citeauthor{rane2018sentiment}&3 & 14640&71.76\%&71.89\%&\textbf{72.44\%}&\textbf{72.44\% (5\%)}
\\
\verb|CM| & \citeauthor{collins2018evolutionary} &4 &5218&\textbf{93.73\%}&92.92\%&87.04\%&\textbf{93.73\% (0\%)}
\\
\verb|CB| & \citeauthor{uzzi2016four}&4 &32000 &99.98\%&\textbf{99.99\%}	&\textbf{99.99\%}&\textbf{99.99\% (1\%)}\\ 
% \verb|MR| &72.61&75.41	&81.74\\ 
\verb|HS|  & \citeauthor{davidson2017automated}&3 & 20941 &69.06\%&68.65\%&\textbf{69.95\% }&\textbf{69.95\% (5\%)}
\\ 
\verb|MR|   & \citeauthor{pang2002thumbs}&2 &2000&72.61\%&75.41\%	&76.22\%&\textbf{81.74\% (3\%)}\\ 
\verb|POL|&  \citeauthor{Pang+Lee+Vaithyanathan:02a}&2 & 1400 &70.70\%&\textbf{71.56\%}&70.5\%&\textbf{71.56\% (1\%)}\\
\verb|PSC|&  \citeauthor{collins2018evolutionary}& 5 & 3117 &59.05\%&59.18\%&61.63\%& \textbf{68.17\% (3\%)}
\\
\verb|QC|& \citeauthor{li2002learning}&	6  & 5142 &	86.88\%&86.51\%&\textbf{87.00\%}&\textbf{87.00\% (5\%)}
\\ 
\verb|RS|  & \citeauthor{kotzias2015group} &2& 3000&79.82\%&\textbf{80.82\%}&78.64\%	&\textbf{80.82\% (1\%)}
\\
\verb|SE| & \citeauthor{nakov2019semeval} & 3&13231 &57.68\%&56.56\%&58.85\%&\textbf{58.95\% (10\%)}
\\
\verb|SMSS| & \citeauthor{almeida2011contributions} &2& 9416	&\textbf{97.78\%}&	\textbf{97.78\%}	&96.81\%&\textbf{97.78\% (1\%)}
\\
\verb|YTS|  & \citeauthor{alberto2015tubespam} & 2& 1948 &94.78\%	&\textbf{96.64\%}&	95.11\%&\textbf{96.64\% (1\%)}
\\

\verb|20NG|  & \citeauthor{adi2014classification} &20&20000&59.11\% &57.59\%	&59.21\%&\textbf{60.26\% (3\%)}
\\
\hline
\end{tabular}
\caption{Performance comparison of trained LSTM models when top $k\%$ difficult samples are removed from the training set. $F1_{Best}$ represents the best score observed for $k \in [0,1,3,5,10,20]$. $F1_{0\%}$ represents baseline performance when no samples are removed.}
\label{tab:results}
\end{table*}

\subsection{Experiment Setup}
\label{ssec:exp_setup}

\subsubsection{Data Preprocessing}
For all the experiments, standard pre-processing steps such as removal of special characters, stop-words, conversion to lowercase, tokenization, etc. have been performed. For generating text representation, we consider two encoding strategies - (i) Average pre-trained word embeddings for tokens in input text using Word2Vec \cite{mikolov2013efficient} and (ii) intermediate layer representation of trained Long Short Term Memory (LSTM) model \cite{hochreiter1997long}. For computing penalty scores using sigmoid functions we use $a=5$ and $b=10$. 

\subsubsection{Evaluation Strategy}
For each dataset, we identify the top $k\%$ difficult samples as described in Sec \ref{ssec:identify_difficult} from the training set for $k \in [0, 1, 3, 5, 10, 20]$. We train two sets of classifiers, one with the complete training set and second after filtering the training set of the difficult samples using different values of $k$ and compare the macro F1 scores for both sets of classifiers on held out test sets. For text representations generated using Word2Vec, we utilize an SVM model with RBF kernel with hyperparameters $C$ \& $\gamma \in [0.001, 0.01, 0.1, 1, 10, 100]$ that are tuned using grid search on the validation set. The LSTM model is trained with the embedding layer initialized with one-hot vector representation of the input text where the maximum vocabulary size is $V_{max} = 10000$ and the maximum sequence length is $S_{max} = 250$ and the network parameters are tuned on the validation set.

\section{Results and Discussion}
\label{sec:results_discussion}
% Table \ref{tab:results} shows the results of F1 scores obtained on the datasets by removing the top $k\%$ difficult samples from the training set.

%\subsection{Effect of Removing Difficult Samples}
\subsection{Analysis of Removing Difficult Samples}
\label{ssec:erds}
As seen from the results shown in Table \ref{tab:results}, an improvement in the F1 scores (upto $\sim$9\%) is seen for most of the datasets. The values for $F1_{Best}$ indicate that the improvement is generally observed for $k \in [1,3,5]$. Although we saw a consistent improvement in the performance of the trained models on the updated train set (difficult samples removed) for lower values of $k$, we observed a consistent drop in performance when top 10\% or 20\% difficult samples were removed. This suggests that the top-ranked difficult samples in the training set ($k \in [1,3,5]$) help in model generalization while for $k>5$ the trained models seem to overfit on the training set resulting in poor performance on the test set. The \verb|CM| dataset \cite{collins2018evolutionary} is an exception to this general trend. On further investigation, we observed that the top ranked difficult samples belong to a minority class and removing them hurts the model performance for test samples from that class. Thus, for text datasets, we observe that semantically difficult samples pose a challenge for the downstream model and removing them in most cases improves its performance.

\begin{figure*}[t]
\centering\includegraphics[width=1.0\linewidth]{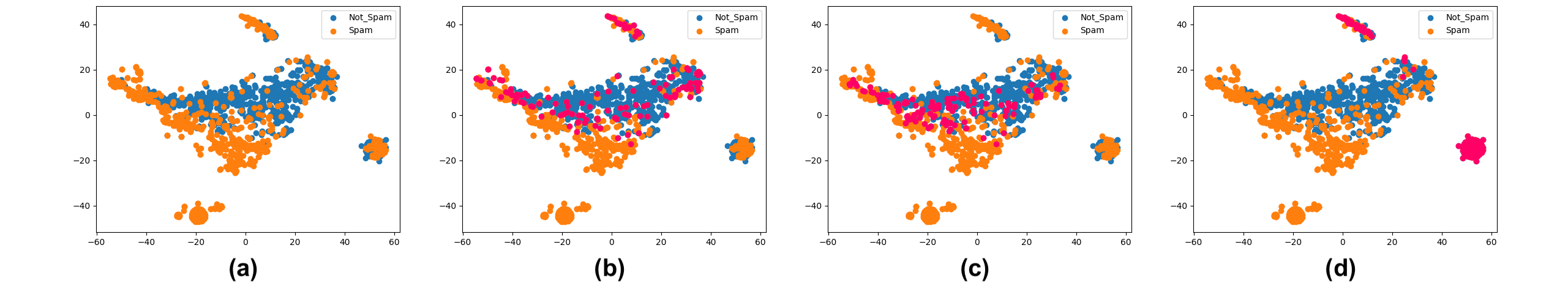}
  \caption{(a) Original TSNE plot for YTS (b) Both cases - top 10\% difficult samples (c) Case-1 only - top 10\% difficult samples (d) Case-2 only - top 10\% difficult samples (Difficult samples are marked in \textcolor{custom_pink}{\textbf{pink}})} 
  \label{fig:yts_cases}
\end{figure*}

% \begin{table*}[t]
% \centering
% \begin{tabular}{ccccccc}
% \hline
% \hline
% \textbf{$Dataset$} & \textbf{$F1_{0\%}$} &  \textbf{$F1_{1\%}$} & \textbf{$F1_{3\%}$} &  \textbf{$F1_{5\%}$} & \textbf{$F1_{10\%}$} & \textbf{$F1_{20\%}$} \\
% \hline
% \hline
% \verb|YTS| & & & & & & 
% \\
% \hline
% \end{tabular}
% \caption{Performance of trained LSTM model corresponding to $F1_{Best}$ when top $k\%$ difficult samples are removed from the validation set.}
% \label{tab:val_results}
% \end{table*}

% \begin{table}[b]
% \centering
% \begin{tabular}{p{2cm}p{5cm}}
% \hline
% \textbf{$Class$} & \textbf{$Original \; Sample$}
% \\
% \hline
% \hline
% \verb|Spam|& He gets more views but has less subscribers lol
% \\
% \hline
% \verb|Not Spam| & We get it, you came here for the views... Ôªø
% \\
% \hline
% \hline
% \end{tabular}
% \caption{Difficult samples in YTS}
% \label{tab:yts_samples}
% \end{table}

\begin{table}[b]
\centering
\begin{tabular}{p{2cm}p{5cm}}
\hline
\textbf{$Class$} & \textbf{$Original \; Sample$}
\\
\hline
\hline
\verb|Spam|&He gets more views but has less subscribers lol
\\
\hline
\verb|Spam| &Yea stil the best WK song ever<br />Thumbs up of you think the same<br />ï»¿
\\ 
\hline
% \verb|Not Spam| & Hello Brazil ðŸ˜»âœŒðŸ’“ðŸ˜»ðŸ‘ï»¿
\verb|Not Spam| & Hello Brazil ðŸ˜»âœŒðŸ’“ðŸ˜
\\
\hline
% \verb|Not Spam| & P E A C E  \&amp;  L O V E  ! !ï»¿
\verb|Not Spam| & We get it, you came here for the views... Ôªø
\\ 
\hline
\end{tabular}
\caption{Top ranked difficult samples in YTS using our approach}
\label{tab:yts_samples}
\end{table}

%\subsection{Regions of Difficult Samples}
\subsection{Analysis of Difficult Samples}
\label{ssec:dscds}

\begin{figure}[t]
    \centering\includegraphics[width=0.6\linewidth]{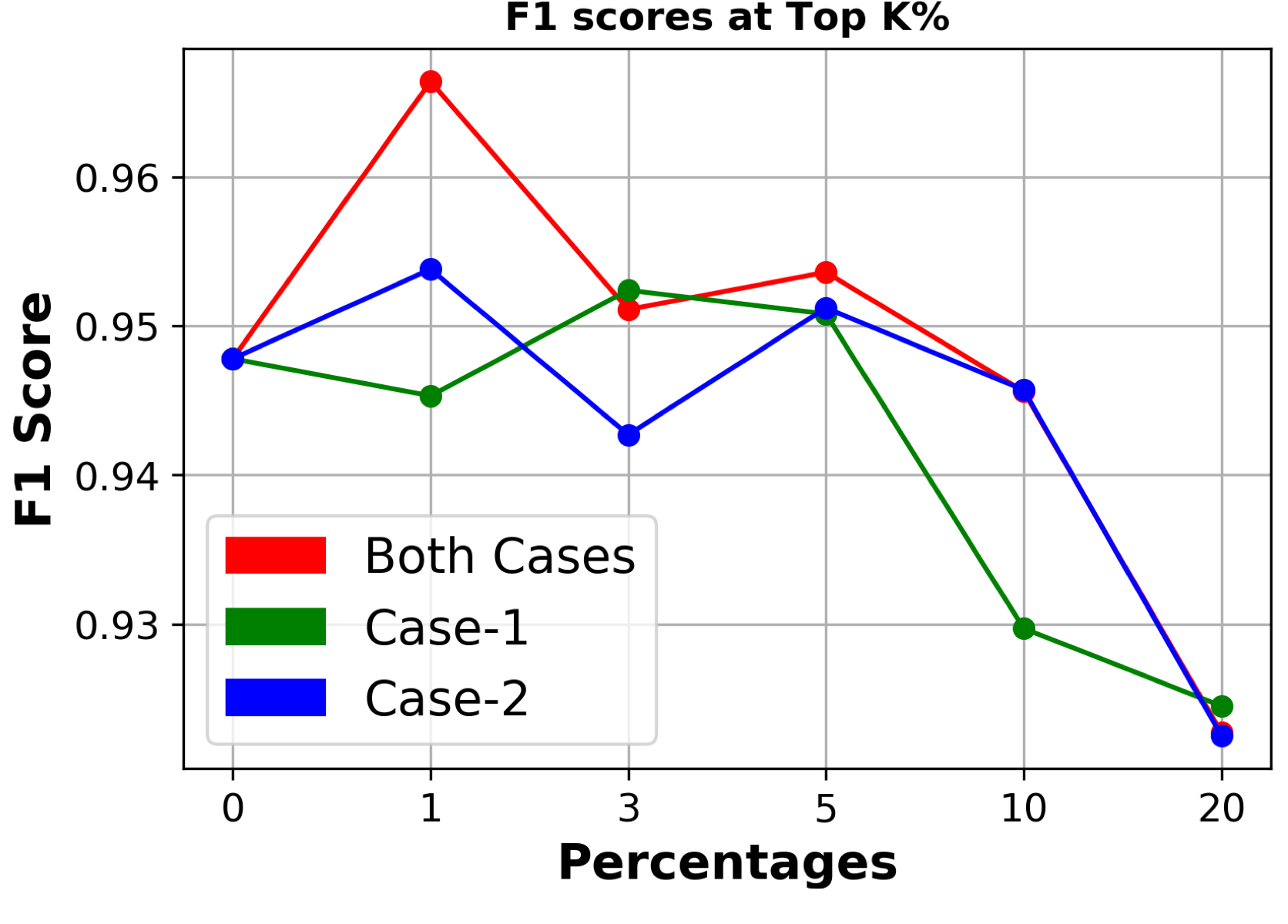}
  \caption{YTS Dataset - Plot of F1 scores for all cases for $k \in [0, 1, 3, 5, 10, 20]$. (Baseline accuracy $F1_{0\%}$ = 0.9478 for all cases).}
  \label{fig:yts_f1}
\end{figure}

Figure \ref{fig:yts_cases} shows the various regions in the TSNE plots corresponding to the top-ranked difficult samples belonging to various cases as discussed in Sec \ref{ssec:difficulty}. The difficult samples due to \textit{case 1} (Figure \ref{fig:yts_cases}(c)) belong to the region of maximum overlap between the class distributions. On the contrary, difficult samples due to \textit{case 2} (Figure \ref{fig:yts_cases}(d)) belong to isolated clusters of samples that are not representative of the respective overall class distributions. Figure \ref{fig:yts_cases}(b) shows the region of difficult samples when both cases are considered for \verb|YTS| dataset. Figure \ref{fig:yts_f1} shows the trend of the F1 scores for all three cases. As observed, the performance gain is higher when difficult samples from both cases are considered than difficult samples from only \textit{case 1} or \textit{case 2} for $k \in [1,3,5]$, while for $k>5$, the performance dips for all configurations.

Table \ref{tab:yts_samples} showcases a few examples of top-ranked difficult samples identified from \verb|YTS| dataset. As seen from examples of \verb|Spam| class, the constituent words are common with the \verb|Not Spam| class and the overall semantic meaning does not specifically indicate the label \verb|Spam|. These particular samples lie in the overlap region as shown in Figure \ref{fig:yts_cases}(b) and removing them from the training set improves the generalizability of the trained model. Similar insights can be derived for the difficult samples belonging to \verb|Not Spam| class.

% \begin{table}[t]
% \centering
% % \begin{tabular}{p{1.4cm}p{1.4cm}p{1.6cm}p{1.4cm}}
% \begin{tabular}{ccccc}
% \hline
% \hline
% \textbf{$Dataset$}& \textbf{$F1^{W2V}_{0\%}$}  & \textbf{$F1^{W2V}_{Best}$} & \textbf{$F1^{LSTM}_{0\%}$}  &  \textbf{$F1^{LSTM}_{Best}$} \\
% \hline
% \hline
% \verb|RS| &53.33\%&58.50 (1\%)&79.82\%&80.82\% (1\%)
% \\
% \verb|SMSS| &94.78\%&95.95\% (1\%)&97.78\%&97.78\% (1\%)
% \\
% \verb|YTS| &84.91\% &86.69 (1\%)\%&94.78\%&96.64\% (1\%)
% \\
% \hline
% \end{tabular}
% \caption{Comparison of performance of various classifier models}
% \label{tab:models}
% \end{table}

\subsection{Effect of Text Encoding and Model}
\label{ssec:eem}

% Finally, we study the effect of employing different text encoding schemes and model architectures for identifying the difficult samples in training set. 
Table \ref{tab:models} shows the F1 scores obtained after removing difficult samples identified using Word2Vec embeddings and SVM classifier as discussed in Sec \ref{ssec:exp_setup}. We observe a similar improvement in the model performance after removing the top 1\% difficult samples as shown in Table \ref{tab:results}. We replicated the experiments to use TF-IDF \cite{ref1} as well as Fasttext \cite{bojanowski2017enriching} embeddings and observed a similar trend. Thus, the identification of difficult samples is agnostic to the underlying text encoding scheme and improves the performance of general text classification model.

% Although, the performance numbers are lower compared to LSTM model as shown in Table \ref{tab:results}.

\begin{table}[t]
\centering
% \begin{tabular}{p{1.4cm}p{1.4cm}p{1.6cm}p{1.4cm}}
\begin{tabular}{ccc}
\hline
\hline
\textbf{$Dataset$}& \textbf{$F1^{W2V}_{0\%}$}  & \textbf{$F1^{W2V}_{Best}$} \\
\hline
\hline
\verb|RS| &53.33\%&58.50\% (1\%)
\\
\verb|SMSS| &94.78\%&95.95\% (1\%)
\\
\verb|YTS| &84.91\% &86.69\% (1\%)
\\
\hline
\end{tabular}
\caption{Performance of trained SVM model using Word2Vec embeddings.}
\label{tab:models}
\end{table}

\subsection{Application to Human In Loop Systems}
\label{ssec:dsdi}

% Difficult samples in dataset could arise due to multiple issues. Some of them can be attributed to the data gathering process which might induce noisy labels. It can also be an intrinsic property of the dataset where gathered labels are not sufficient to capture the overall set of semantic topics and fine-grained labels are necessary. The text encoding scheme also plays an important role in determining if the representations are sufficiently discriminative to train the classifier on. As seen from our analysis, the difficult samples are most likely to get misclassified by the trained model. Thus, for real word ML applications with a human in the loop system, model predictions on only the difficult samples could be verified by the human expert during the training phase to reduce their effort and improve the overall performance.

Difficult samples in a dataset could arise due to the data gathering process which might induce noisy labels. It can also be an intrinsic property of the dataset where gathered labels are not sufficient to capture the overall set of semantic topics and fine-grained labels are necessary. As seen from our analysis, the difficult samples are most likely to get misclassified by the trained model. Thus, our approach can quickly identify the semantically difficult samples in a dataset which a data scientist could use to review their labels with the help of a domain expert or build provisions in the modeling pipeline to address them.

\section{Conclusion and Future work}
We present a method to identify semantically difficult samples and suggest two scenarios of how such samples can affect the training data and the corresponding model. We show by extensive experimental evaluation that the classifiers trained after removing difficult samples show a gain in performance ($\sim$9\%) as compared to the classifiers trained on the full training set. Thus, we show that training data assessment is an important pre-step before training classifier models. A related problem is to automatically identify the optimum value of $k$ for a dataset and remediate the data of the difficult samples, which we plan to explore in the future.

%affect a model, and suggest a method to detect data samples using a cumulative penalty method.  and show by experimental evaluation on $12$ datasets, the effect of cleaning the datasets by 
%The assessment of the quality of the training data is something that is not talked about much. However, it can be observed that by improving the quality of training data, the performance improves for the very same method one intends to use for classification. Simply finding the difficult samples and removing them leads the classifier to perform better. The results clearly convey that performance of the classifier is dataset specific. For some datasets, removing some difficult samples, the performance improves to a great extent. While for some datasets, which are cleaner, the samples who have a relatively high penalty amongst other samples in the same dataset, might not be actually difficult samples. Thus, depending on the dataset, decision on the percentage of samples to be removed should be made. As seen in the TSNE plots for YTS dataset, the samples that have been highlighted, are the samples which might be causing difficulty for the classifier to perform. Thus, using the characteristics of the dataset, the best possible method of removing the samples based on the provided definition of the penalty has to be decided.

\bibliographystyle{acl_natbib}
\bibliography{refs}

\end{document}